\documentclass[lettersize,journal]{IEEEtran}
\usepackage{amsmath,amsfonts}
\usepackage{algorithmic}
\usepackage{algorithm}
\usepackage{array}
\usepackage[caption=false,font=normalsize,labelfont=sf,textfont=sf]{subfig}
\usepackage{textcomp}
\usepackage{stfloats}
\usepackage{url}
\usepackage{verbatim}
\usepackage{graphicx}
\usepackage{amssymb}
\usepackage{multirow}
\usepackage{cite}
\begin{document}

\title{OFD-Net: Teacher-Free Reliable Semi-supervised Medical Image Segmentation with Orthogonal Feature Disentanglement Net of Foreground-Background}

\author{Shao-feng~Jiang, Zhe-yang~Jing\textsuperscript{\dag}, Qin~Lu, Huan-huan~Shi, Zhen~Chen, Cong-xuan~zhang  and Chen~Yi\textsuperscript{\dag}%
\thanks{Shao-feng Jiang, Zhe-yang Jing, Qin Lu, Huan-huan Shi, Zhen Chen, Cong-xuan zhang and Chen Yi are with the Department of Biomedical Engineering, and the Key Laboratory of Nondestructive Testing, Ministry of Education, Nanchang Hangkong University, Nanchang 330063, China.}%
\thanks{\textsuperscript{\dag}Corresponding authors: Zhe-yang Jing (Email: \texttt{1255813468@qq.com}); Chen Yi (Email: \texttt{470835297@qq.com})}%
}


\maketitle

\begin{abstract}
 Semi-supervised learning (SSL) is an effective solution for medical image segmentation with limited annotations. Existing SSL methods mainly rely on pseudo-labels generated by teacher-student supervision or cross-network consistency. However, these methods lack an explicit structural reference for judging pseudo-label quality. Low-quality pseudo-labels may lead to unreliable training, error accumulation and confirmation bias when processing unlabeled data with substantial appearance variations. To address this issue, we proposed OFD-Net, a teacher-free single-network framework for reliable semi-supervised medical image segmentation. OFD-Net employs an Orthogonal Feature Disentanglement Module (OFDM) to capture OFD features for reliable SSL by disentangling unlabeled data into background and foreground representations with a reliable structural distribution, thereby effectively reducing error accumulation and alleviating confirmation bias among unlabeled data. Specifically, OFD-Net explicitly employs a Disentanglement Guidance Module (DGM) to inject the resulting structural priors of foreground-background into the decoder by deformable convolution processing, and outputs predictions with clearer foreground representations. Based on DGM and the OFDM, we further develop a reliability-aware pseudo-label learning mechanism that evaluates unlabeled supervision according to the structural consistency between the main prediction and the disentangled foreground-background responses, and then down-weights unreliable regions during training. Extensive experiments on four public medical image segmentation benchmarks, namely ISIC-2016, Kvasir-SEG, Synapse, and ACDC, validate the effectiveness of OFD-Net. On ACDC, OFD-Net with a SegFormer-B4 backbone attains a Dice score of 90.42\% and a Jaccard score of 86.25\% with only 10\% labeled data, exceeding KnowSAM by 0.86 and 4.59 percentage points, respectively, while achieving an HD95 of 1.36 mm and an ASD of 0.78 mm. These results confirm that orthogonal foreground-background disentanglement enables OFD-Net to establish an efficient and reliable training paradigm within a teacher-free single-network framework.
\end{abstract}


\section{Introduction} \label{sec:intro}

\IEEEPARstart{I}{n} recent years, deep learning models, especially convolutional neural networks (CNNs) and Transformers, have achieved remarkable progress in medical image segmentation \cite{ronneberger2015u, chen2021transunet}. However, the strong performance of fully supervised methods \cite{cao2022swin, 10183842} depends heavily on large-scale, high-quality pixel-level annotations. In clinical practice, such annotations are expensive and time-consuming to obtain. This creates a serious label-scarcity problem. To reduce the reliance on labeled data, semi-supervised learning (SSL) has become an important solution for medical image segmentation \cite{chen2021semi, tarvainen2017mean, yao2022enhancing}. Its goal is to learn robust feature representations from a small labeled set and a large unlabeled set.

Most existing SSL methods follow the consistency regularization paradigm. Representative examples include Mean Teacher \cite{tarvainen2017mean} and multi-view co-training methods \cite{wu2021semi}. Although these methods have achieved promising results, they still suffer from confirmation bias and, more importantly, carry an inherent risk of introducing unreliable pseudo-labels into training. Specifically, incorrect predictions on difficult unlabeled samples produce noisy supervision, and these errors keep accumulating during training. To alleviate this problem, many existing methods introduce multi-stream or multi-teacher architectures, together with diverse consistency constraints, to improve robustness. These strategies can improve stability through output alignment \cite{Wang_2023_CVPR}. However, they still focus mainly on prediction-level or network-level constraints, rather than explicitly modeling the internal feature representation. Therefore, when unlabeled data exhibit large appearance variations due to imaging conditions, noise, artifacts, or individual differences, these methods remain susceptible to unreliable pseudo-labels and confirmation bias during training \cite{9207304}.

\begin{figure}[!t]
\centering
\includegraphics[width=\columnwidth]{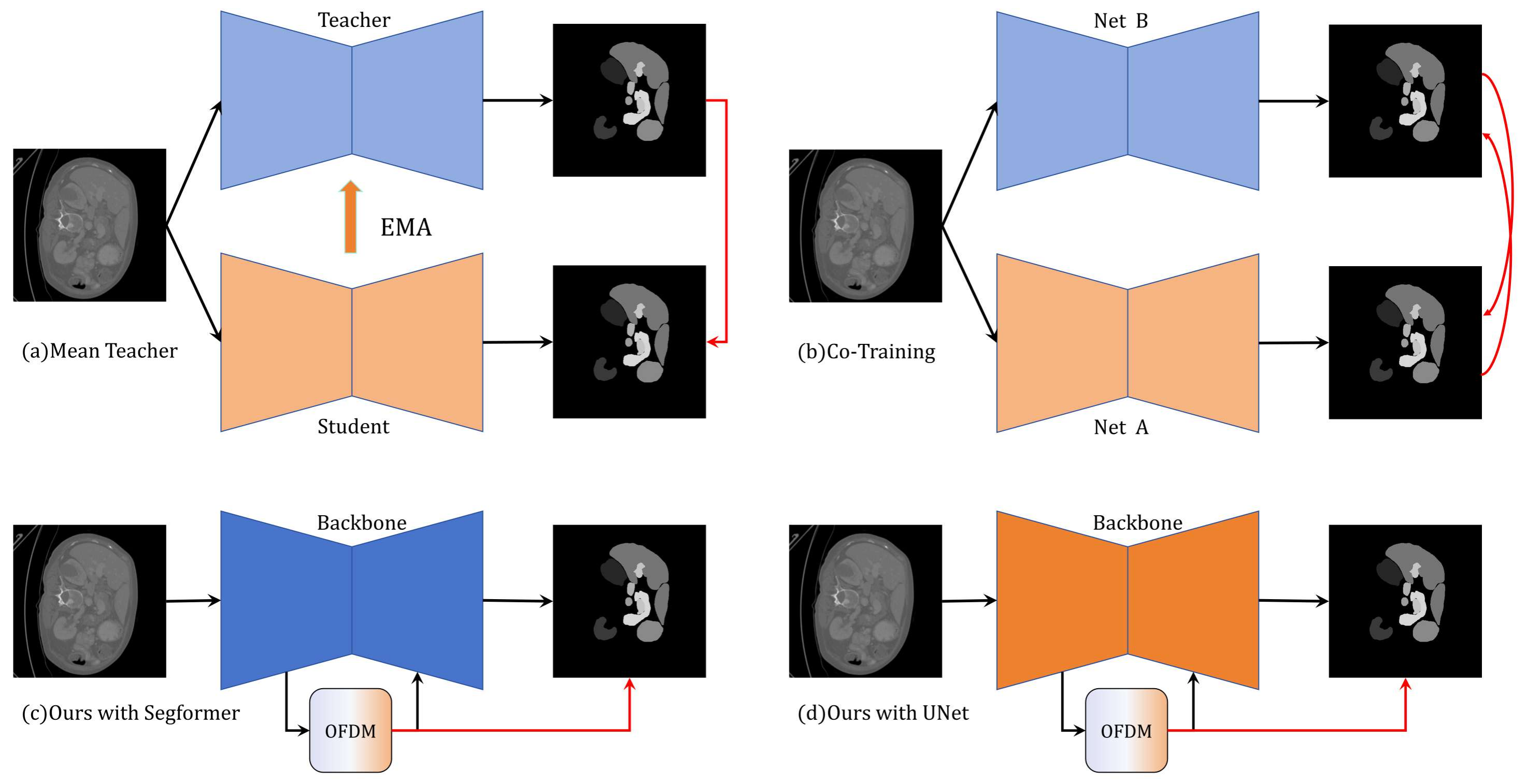}
\caption{Conceptual comparison between conventional semi-supervised segmentation frameworks and the proposed OFD-Net. Traditional methods, such as Mean Teacher and Co-Training, mainly rely on supervision transfer across multiple networks. In contrast, OFD-Net builds an internal structural reference within a single backbone through the Orthogonal Feature Disentanglement Module (OFDM), and uses it to guide prediction refinement. The same design can be instantiated on different backbones, as illustrated with SegFormer and U-Net.}
\label{fig_1}
\end{figure}

This issue is closely related to feature-level semantic entanglement. Existing studies show that medical images follow shared anatomical and biological principles, so organs and lesions often have similar topological patterns. At the same time, different subjects and acquisition conditions introduce obvious appearance variations \cite{CHARTSIAS2019101535, dou2019pnp}. Under limited supervision, this combination of structural regularity and appearance variation makes it difficult for the model to separate foreground from background in the feature space, especially in regions with blurry boundaries or strong noise. As a result, background responses may leak into foreground representations, giving rise to ambiguous predictions or even high-confidence yet structurally incorrect predictions on unlabeled samples. Such predictions may be mistakenly treated as reliable pseudo-labels in conventional consistency-based frameworks. Therefore, pseudo-label reliability cannot be determined solely by confidence, because the prediction itself may already arise from entangled and structurally unreliable features \cite{CHARTSIAS2019101535}.

Besides these architectural and representation-level challenges, existing semi-supervised methods do not fully exploit the unique role of labeled data in unlabeled learning. Labeled samples provide precise anatomical structure and topological priors. However, many existing methods \cite{HE2025103626, Bai_2023_CVPR} use them only as direct supervision or combine them with unlabeled samples through simple mixing strategies. They do not fully use labeled data as a structural semantic reference. Without such a reliable reference, the model often relies on image-level consistency, confidence filtering, or statistical heuristics to screen pseudo-labels. These strategies, however, cannot explicitly judge whether a prediction is structurally reasonable. As a result, the optimization process may still favor pseudo-labels that have high confidence but wrong semantics. This leads to an optimization trap \cite{9207304}. It also causes the reliable structural knowledge learned from labeled data to be gradually overwhelmed by appearance bias from abundant unlabeled samples \cite{NEURIPS2018_c1fea270, liu2025confidence}.

To address these issues, we propose OFD-Net, a teacher-free single-network framework for reliable semi-supervised medical image segmentation, driven by orthogonal foreground-background disentanglement. Unlike conventional methods that rely on external teacher networks or cross-network consistency, OFD-Net introduces an Orthogonal Feature Disentanglement Module (OFDM) at the encoder bottleneck. OFDM explicitly decomposes deep features into foreground and background representations, and thus establishes a clear foreground-background structural reference in the latent space. Based on this design, we further introduce a Disentanglement Guidance Module (DGM) to inject this structural prior into the decoder and improve the discriminative ability of the main segmentation branch. In addition, based on the foreground-background responses from the disentangled branches, we develop a reliability-aware pseudo-label learning mechanism. Instead of relying only on confidence, this mechanism evaluates the trustworthiness of unlabeled supervision from the perspective of structural consistency. It then performs region-wise reweighting of pseudo-label supervision, which helps suppress error propagation from unreliable regions. As shown in Fig.~\ref{fig_1}, OFD-Net builds an internal structural reference within a single backbone and uses it to guide both decoding and pseudo-label learning. This design distinguishes OFD-Net from conventional semi-supervised frameworks that depend on supervision transfer across networks.

The main contributions of this paper are summarized as follows:
\begin{itemize}
    \item[(1)] We propose OFD-Net, a teacher-free single-network framework for reliable semi-supervised medical image segmentation. The proposed framework improves pseudo-label reliability from the perspective of foreground-background feature disentanglement, rather than relying only on prediction-level or cross-network consistency.
    \item[(2)] We design an Orthogonal Feature Disentanglement Module (OFDM) and a Disentanglement Guidance Module (DGM). OFDM establishes a clear foreground-background structural reference in the latent space, and DGM injects this prior into the decoder to improve segmentation discrimination.
    \item[(3)] We develop a reliability-aware pseudo-label learning mechanism that transfers the structural criterion learned from labeled data to unlabeled samples and enables region-wise supervision reweighting based on structural consistency and separability.
    \item[(4)] Experiments on multiple public medical image segmentation benchmarks show that the proposed method achieves clear advantages under limited annotations and demonstrates strong effectiveness and robustness.
\end{itemize}

\section{Related Work} \label{sec:related}

\subsection{Semi-supervised Medical Image Segmentation}

Semi-supervised medical image segmentation aims to exploit limited labeled data together with abundant unlabeled data for joint learning. A large body of existing methods follows the consistency-regularization paradigm. Mean Teacher \cite{tarvainen2017mean} introduces an EMA-updated teacher to supervise the student network on unlabeled data. CCT \cite{Ouali_2020_CVPR} enforces prediction consistency between the main decoder and multiple auxiliary decoders under different perturbations. DTC \cite{Luo_Chen_Song_Wang_2021} and SLCNet \cite{10.1007/978-3-031-16452-1_14} further incorporate auxiliary geometric tasks, such as signed distance field prediction or location-aware constraints, to improve boundary modeling. URPC \cite{10.1007/978-3-030-87196-3_30} exploits pyramid consistency across multiple resolutions, while SASSNet \cite{10.1007/978-3-030-59710-8_54} imposes shape-aware constraints to encourage anatomically plausible predictions. More recently, ST++ \cite{Yang_2022_CVPR} showed that strong perturbations can substantially improve pseudo-label learning, and AdaptFRCNet \cite{HE2025103626} further introduced frequency-domain consistency to extract style-invariant features. In addition, recent SAM-based semi-supervised methods have begun to explore the use of foundation models in this setting. For example, KnowSAM \cite{huang2025learnable} introduces a learnable prompting SAM-induced knowledge distillation framework that combines multi-view co-training with SAM-guided knowledge transfer, thereby improving robustness under limited annotations.

Despite their strong performance, these methods mainly enhance semi-supervised learning through teacher-student supervision, multi-branch consistency, auxiliary-task regularization, perturbation-based prediction alignment, or external knowledge transfer. Their core supervision still acts primarily at the output level or across networks. As a result, when unlabeled samples exhibit substantial appearance variation, the learned predictions may remain vulnerable to feature-level foreground-background entanglement. In contrast, our method does not rely on external teachers, cross-network consistency, or SAM-induced supervision. Instead, it establishes an internal foreground-background structural reference within a teacher-free single-network framework and uses it to support both decoding and reliable pseudo-label learning.

\subsection{Feature Disentanglement and Representation Structuring}

Feature disentanglement aims to separate entangled representations into more interpretable and independent factors, and has shown strong potential in medical image analysis. SDNet \cite{CHARTSIAS2019101535} is an early representative that disentangles anatomical structure from imaging appearance to improve robustness in cross-modality analysis. In semi-supervised settings, related methods have also attempted to reduce the mismatch between labeled and unlabeled representations from different perspectives. BCP \cite{Bai_2023_CVPR} alleviates distribution discrepancy through bidirectional copy-paste and pixel-level mixing. AllSpark \cite{wang2024allspark} uses cross-attention to align unlabeled features with labeled information. SKCDF \cite{Zhang_2025_CVPR} further introduces complementary semantic knowledge to reduce interference from unlabeled data.

However, most of these methods mainly rely on data mixing, feature interaction, or implicit alignment. They do not explicitly impose geometric constraints on the latent space to construct a clear foreground-background structural reference. Consequently, when supervision is limited, it remains difficult to guarantee that foreground and background cues are cleanly separated in the learned representation. Different from these methods, our approach explicitly disentangles deep features into mutually exclusive foreground and background branches through an orthogonality constraint, and then uses this structured representation not only for feature learning but also for decoder guidance and pseudo-label verification.

\subsection{Reliability-aware Learning for Pseudo-label Supervision}

Because pseudo-labels inevitably contain noise, reliability estimation plays a central role in semi-supervised learning. UA-MT \cite{10.1007/978-3-030-32245-8_67} uses prediction entropy to estimate uncertainty and masks out unreliable regions during teacher-student learning. DAW \cite{NEURIPS2023_c28ef844} and ARCO \cite{NEURIPS2023_1f7e6d5c} further refine this idea by introducing distance-aware or attention-based weighting strategies to reduce the effect of noisy boundaries. Another line of work explores optimization-based reweighting. MPL \cite{Pham_2021_CVPR} and MetaSSL \cite{11150469} use validation feedback from labeled data to adjust pseudo-label weights through meta-learning. In addition, confidence-based methods such as UPS \cite{Rizve2021InDO} and FixMatch \cite{NEURIPS2020_06964dce} retain only high-confidence predictions for unlabeled supervision.

Although effective, these methods mainly estimate reliability through uncertainty statistics, confidence thresholds, or validation-driven reweighting. Such signals do not explicitly verify whether a pseudo-label is structurally reasonable. In particular, a prediction can still be confident while remaining semantically incorrect. Our method addresses this limitation by introducing a disentanglement-based reliability criterion. Instead of judging pseudo-label quality solely from confidence, we evaluate whether the main prediction is consistent with the disentangled foreground branch and sufficiently separable from the background branch. This design allows labeled data to provide not only direct supervision, but also a transferable structural criterion for verifying unlabeled predictions.

\section{Methods}
\textbf{Definitions and Notations:} For semi-supervised learning, the training set consists of a limited labeled subset $D_L = \{(x_n^l, y_n)\}_{n=1}^{N_L}$ and an unlabeled subset $D_U = \{x_m^u\}_{m=1}^{N_U}$, where $N_L$ and $N_U$ denote the numbers of labeled and unlabeled samples, respectively, and $N_L \ll N_U$. The full training set is denoted by $D = D_L \cup D_U$. Here, $x \in \mathbb{R}^{H \times W \times C}$ denotes an input image, where $H$ and $W$ are the image height and width, and $C$ is the number of input channels. $y$ denotes the corresponding pixel-wise annotations. We use $n$ to index labeled samples and $m$ to index unlabeled samples.

\begin{figure*}[!t]
\centering
\includegraphics[width=\textwidth]{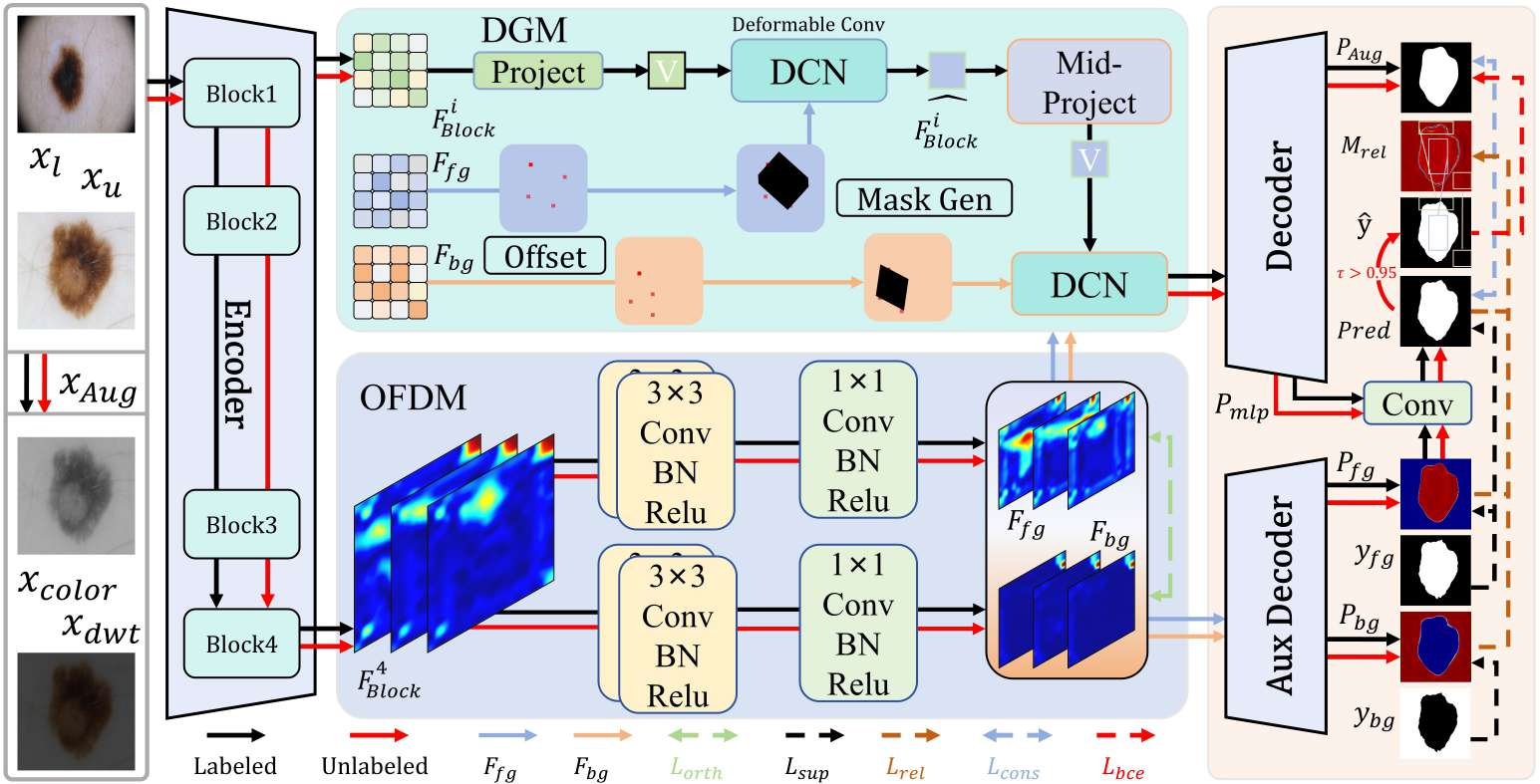}
\caption{Overall architecture of the proposed single-network OFD-Net, built on a SegFormer-B4 backbone with orthogonal foreground-background disentanglement and reliability-aware semi-supervised learning.}
\label{fig_2}
\end{figure*}

\subsection{Overall Architecture}\label{sec:architecture}

As shown in Fig.~\ref{fig_2}, OFD-Net is a teacher-free single-network framework for semi-supervised segmentation, built on a SegFormer-B4 backbone. Different from conventional teacher-student or cross-network consistency methods, the proposed framework does not rely on external supervisory branches to assess unlabeled predictions. Instead, it explicitly constructs an internal foreground-background structural reference and uses this reference to support both decoder refinement and pseudo-label reliability estimation. In this way, OFD-Net replaces external supervision transfer with internally learned, structure-aware reliability modeling.

Given an input image $x$, the encoder extracts multi-scale feature maps $\{F_{Block}^i\}_{i=1}^4$, where $F_{Block}^i \in \mathbb{R}^{C_i \times H_i \times W_i}$ denotes the feature tensor produced by the $i$-th backbone stage. The deepest encoder feature $F_{Block}^4$ is then fed into an Orthogonal Feature Disentanglement Module (OFDM), which decomposes it into foreground features $F_{fg}$ and complementary background features $F_{bg}$. These disentangled features serve two main purposes. First, an auxiliary decoder maps $F_{fg}$ and $F_{bg}$ into foreground and background responses, denoted by $P_{fg}$ and $P_{bg}$, respectively. These responses provide an explicit foreground-background structural reference for subsequent reliability estimation. Second, the disentangled foreground-background features are injected into the main decoder through the Disentanglement Guidance Module (DGM), which guides multi-scale feature fusion and improves foreground-background discrimination.

Training proceeds in two stages. In both stages, we introduce semantic-frequency dual perturbations to improve robustness to appearance variation and frequency disturbance, so that the model learns stable foreground-background representations under heterogeneous visual conditions. In Stage I, the model uses only labeled data $D_L$. For each labeled image $x_l$, the network first produces a prediction Pred, which is directly supervised by the ground-truth label $y$. At the same time, the model constructs a perturbed image $x_{Aug}$ with the proposed semantic-frequency perturbations and feeds it into the same network to obtain the corresponding prediction $P_{Aug}$. We then impose a consistency loss between $Pred$ and $P_{Aug}$. This step encourages the network to learn a stable foreground-background disentangled representation that remains robust to appearance variation and frequency disturbance. With labeled supervision, the model also learns a clear and mutually exclusive foreground-background structure. In Stage II, the model further introduces unlabeled data $D_U$ for semi-supervised optimization. For each unlabeled image $x_u$, the network first produces a prediction Pred. We then generate a hard pseudo-label $\hat{y}$ from Pred, retain only high-confidence pixels by bidirectional confidence filtering that retains both high-confidence foreground pixels and high-confidence background pixels, and further modulate the retained regions using the reliability map $M_{rel}$. The corresponding perturbed image $x_{Aug}$ produces the prediction $P_{Aug}$, which is optimized under this filtered pseudo-label supervision. In this way, OFD-Net achieves reliability-aware pseudo-label supervision by ensuring that unreliable or structurally inconsistent regions contribute less to the optimization process.

\subsection{Orthogonal Feature Disentanglement Module (OFDM)}\label{sec:ofdm}

To provide a clear foreground-background reference for subsequent decoding and pseudo-label reliability estimation, we introduce an Orthogonal Feature Disentanglement Module (OFDM) at the encoder bottleneck. This module explicitly separates the deepest feature $F_{Block}^4$ into a foreground-oriented feature $F_{fg}$ and a background-oriented feature $F_{bg}$. Under the proposed teacher-free single-network framework, this disentangled representation serves as the foundation of the subsequent reliability-aware learning process: it not only provides a clearer structural basis for segmentation, but also enables the model to judge whether unlabeled predictions are consistent with the learned foreground-background structure, rather than relying on the indirect regularization typically provided by auxiliary branches or teacher networks in conventional semi-supervised architectures.

Specifically, OFDM contains two parallel projection functions, $\varphi_{fg}(\cdot)$ and $\varphi_{bg}(\cdot)$, which are used to extract the foreground-oriented feature $F_{fg}$ and the background-oriented feature $F_{bg}$, respectively:
\begin{equation}
F_{fg} = \varphi_{fg}(F_{Block}^4)
\end{equation}
\begin{equation}
F_{bg} = \varphi_{bg}(F_{Block}^4)
\end{equation}

As illustrated in Fig.~\ref{fig_2}, the two branches share the same projection architecture. Each branch is composed of two stacked $3 \times 3$ Conv-BN-ReLU blocks followed by a $1 \times 1$ Conv-BN-ReLU block. Since the two branches are structurally symmetric but independently parameterized, they can learn complementary responses from the same encoder feature. To prevent the two branches from learning highly overlapping or entangled representations, $F_{fg}$ and $F_{bg}$ are further optimized with an orthogonality loss, which is formulated as:
\begin{equation}
L_{orth} = \frac{1}{H_b W_b} \sum_{h,w} \frac{\left| F_{fg}(h,w) \cdot F_{bg}(h,w)^T \right|}{\left\| F_{fg}(h,w) \right\|_2 \left\| F_{bg}(h,w) \right\|_2}
\end{equation}

Here, $H_b$ and $W_b$ denote the height and width of the bottleneck feature map, respectively. $F_{fg}(h,w)$ and $F_{bg}(h,w)$ denote the foreground-oriented and background-oriented feature vectors at spatial location $(h, w)$, respectively. $(\cdot)^T$ denotes transpose, $\| \cdot \|_2$ denotes the L2 norm. This constraint minimizes the absolute cosine similarity between $F_{fg}$ and $F_{bg}$ at each spatial location, thereby reducing redundant coupling between the two branches and encouraging them to encode complementary structural cues.

For multi-class segmentation, instead of explicitly imposing $O(K^2)$ pairwise orthogonality constraints among all $K$ classes, we adopt a class-wise one-vs-rest (OVR) on top of a shared encoder. Specifically, for each class $k$, we construct a positive branch $F_{pos}^k$ to model class $k$, and a complementary branch $F_{comp}^k$ to aggregate the remaining $K-1$ classes together with the true background:
\begin{equation}
F_{pos}^k = \varphi_{pos}^k(F_{Block}^4)
\end{equation}
\begin{equation}
F_{comp}^k = \varphi_{comp}^k(F_{Block}^4)
\end{equation}

The corresponding orthogonality constraint is then applied independently to each class-specific branch pair:
\begin{equation}
L_{orth}^k = \frac{1}{H_b W_b} \sum_{h,w} \frac{\left| F_{pos}^k(h,w) F_{comp}^k(h,w)^T \right|}{\left\| F_{pos}^k(h,w) \right\|_2 \left\| F_{comp}^k(h,w) \right\|_2}
\end{equation}

In this way, each class branch preserves the same core idea as in the binary case, namely, establishing a class-specific structural reference through disentanglement between the positive branch and the complementary branch, while avoiding over-constraining semantically related categories through exhaustive pairwise orthogonality across all classes.

\subsection{Disentanglement Guidance Module (DGM)}

To propagate the disentangled foreground-background structure learned by OFDM into the decoder, we introduce a Disentanglement Guidance Module (DGM) along the decoding pathway. DGM uses the disentangled foreground and background features as structural prototypes to guide multi-scale decoding. Specifically, the foreground prototype emphasizes target-related responses, while the background prototype suppresses irrelevant or confusing regions. In this way, the decoder is encouraged to preserve structural consistency with the disentangled representation, rather than relying only on the original backbone features.

For each decoding stage $i$, we first map the disentangled foreground and background features to the corresponding scale, denoted by $F_{fg}^i$ and $F_{bg}^i$, respectively. DGM then uses these two structural priors to modulate the encoder-decoder feature interaction through deformable guidance, producing the guided feature $F_{out}^i$. The decoder receives the guided features $\{F_{out}^i\}$ from different stages and produces $P_{mlp}$. To preserve the structural guidance provided by the disentanglement branch, we further fuse the main decoder output with the foreground prediction branch:
\begin{equation}
\text{Pred} = \text{Conv}(P_{mlp} \oplus P_{fg})
\end{equation}
where $\oplus$ denotes channel-wise fusion. For the perturbed input branch, the network follows the same DGM-guided decoding process, and the corresponding output is denoted as $P_{Aug}$.

\subsection{Reliability Estimation}\label{sec:reliability}

Based on the disentangled representation constructed by OFDM and the structurally guided main prediction produced by DGM, we further develop a reliability-aware pseudo-label learning mechanism to evaluate unlabeled supervision according to structural consistency and separability. In medical images, ambiguous boundaries, background interference, and appearance heterogeneity can still lead to high-confidence yet structurally inconsistent predictions. Therefore, confidence filtering alone cannot fully exclude unreliable pseudo-labels and may even cause error accumulation during training. To address this, we leverage the disentangled foreground-background representation learned by OFDM and construct a reliability map to assess the trustworthiness of pseudo-label supervision. In Stage I, labeled supervision teaches the network to judge the reliability of structural predictions according to foreground-background consistency and separability. In Stage II, the model transfers this learned reliability criterion to unlabeled samples and uses it to spatially weight pseudo-label supervision, thereby reducing error accumulation caused by unreliable pseudo-labels.

To assign explicit foreground and background semantics to the disentangled features, we project $F_{fg}$ and $F_{bg}$ into the probability space through two auxiliary heads $h_{fg}(\cdot)$ and $h_{bg}(\cdot)$, which correspond to the auxiliary decoder in Fig.~\ref{fig_2}:
\begin{equation}
P_{bg} = \sigma\left(h_{bg}(F_{bg})\right)
\end{equation}
\begin{equation}
P_{fg} = \sigma\left(h_{fg}(F_{fg})\right)
\end{equation}

The auxiliary decoder explicitly associates the two disentangled branches with foreground and background probability responses, rather than leaving them as latent feature branches only. Based on the disentangled foreground-background responses, we define the reliability map as:
\begin{equation}
M_{rel} = \left( 1 - \left| \text{Pred} - P_{fg} \right| \right) \cdot \left| P_{fg} - P_{bg} \right|
\end{equation}

This formulation reflects two complementary criteria. The first term measures the consistency between the main prediction Pred and the disentangled foreground response $P_{fg}$. The second term measures the separability between the foreground response $P_{fg}$ and the background response $P_{bg}$. A region is therefore reliable only when Pred agrees with the foreground branch and the foreground-background responses remain sufficiently exclusive.

In Stage II, for each unlabeled image $x_u$, the network first produces a prediction Pred. We then generate the hard pseudo-label $\hat{y}$ from Pred and retain only high-confidence pixels by bidirectional confidence filtering. The corresponding perturbed image $x_{Aug}$ produces the prediction $P_{Aug}$, which is optimized under this filtered pseudo-label supervision. The retained regions are further modulated by the reliability map $M_{rel}$, yielding the final pixel-wise weight for unlabeled supervision.

For binary segmentation, the hard pseudo-label is generated from the main prediction as $\hat{y} = 1(\text{Pred} > 0.5)$. To avoid supervising uncertain pixels, we adopt bidirectional confidence. The retained regions are further modulated by the reliability map $M_{rel}$. Therefore, the final pixel-wise weight for unlabeled supervision is defined as
\begin{equation}
W = 1(\text{Pred} > \tau \text{ or } \text{Pred} < 1 - \tau) \odot M_{rel}
\end{equation}
where $\odot$ denotes element-wise multiplication. The perturbed image $x_{Aug}$ produces the prediction $P_{Aug}$, which is optimized under the filtered and reliability-weighted pseudo-label supervision:
\begin{equation}
L_{sup}^{main} = \frac{\sum \left[ W \odot L_{bce}(P_{Aug}, \hat{y}) \right]}{\sum W + \varepsilon}
\end{equation}

In addition, we impose the same reliability-weighted pseudo-label supervision on the disentangled foreground and background branches:
\begin{equation}
L_{sup}^{aux} = \frac{\sum \left[ W \odot \left( L_{bce}(P_{fg}^{Aug}, \hat{y}_{fg}) + L_{bce}(P_{bg}^{Aug}, \hat{y}_{bg}) \right) \right]}{\sum W + \varepsilon}
\end{equation}

To further encourage reliable predictions, we introduce a reliability maximization term on the predicted reliability map:
\begin{equation}
L_{rel} = \frac{\sum \left[ W \odot L_{bce}(M_{rel}^{Aug}, 1) \right]}{\sum W + \varepsilon}
\end{equation}
where $1$ denotes an all-one target map. The unlabeled objective is therefore defined as
\begin{equation}
L_{unlab} = L_{sup}^{main} + L_{sup}^{aux} + L_{rel}
\end{equation}

Under the class-wise OVR formulation described in Section~\ref{sec:ofdm}, the same reliability principle can be extended to multi-class segmentation. For each class $k$, the class-specific positive and complementary probability responses are defined as:
\begin{equation}
P_{pos}^k = \sigma\left(h_{pos}^k(F_{pos}^k)\right)
\end{equation}
\begin{equation}
P_{comp}^k = \sigma\left(h_{comp}^k(F_{comp}^k)\right)
\end{equation}

The corresponding class-wise reliability term is
\begin{equation}
M_{rel}^k = \left( 1 - \left| \text{Pred}^k - P_{pos}^k \right| \right) \cdot \left| P_{pos}^k - P_{comp}^k \right|
\end{equation}

In this way, each class branch preserves the same reliability logic as in the binary case, namely, a pixel is regarded as reliable only when the main prediction agrees with the class-specific positive branch and the positive and complementary responses remain sufficiently separable. Accordingly, for unlabeled samples in Stage II, pseudo-label generation, confidence filtering, and reliability-weighted supervision can be extended to the multi-class setting in an analogous class-wise OVR manner.

\subsection{Training Strategy and Loss Functions}

Training follows the two-stage scheme described in Section~\ref{sec:architecture}. In both Stage I and Stage II, we introduce semantic-frequency dual perturbations to improve the robustness of foreground-background representation learning under appearance and frequency variations. The semantic perturbation is designed to simulate appearance-level style changes. Concretely, the input image is transformed by strong photometric augmentation, including color and intensity variations. The frequency perturbation is designed to alter the image representation in the frequency domain. Specifically, given an input image $x$, we first apply a discrete wavelet transform (DWT) to decompose it into one low-frequency component and three high-frequency components:
\begin{equation}
(x_{LL}, x_{LH}, x_{HL}, x_{HH}) = \text{DWT}(x)
\end{equation}
where $x_{LL}$ denotes the low-frequency component, and $x_{LH}$, $x_{HL}$ and $x_{HH}$ denote the high-frequency components along different orientations. We then perturb the frequency representation by reweighting the high-frequency components through a learnable transformation $G(\cdot)$, and reconstruct the image by inverse wavelet transform:
\begin{equation}
x_{freq} = \text{IDWT}(x_{LL}, G(x_{LH}, x_{HL}, x_{HH}))
\end{equation}
In this way, redundant texture responses can be suppressed, while boundary-related structural cues are emphasized.

The Stage-I objective is defined as:
\begin{equation}
L_{stage1} = L_{sup} + \lambda_{orth} L_{orth} + \lambda_{cons} L_{cons}
\end{equation}

The supervised loss consists of the main prediction loss and two auxiliary BCE losses on the disentangled foreground/background branches:
\begin{equation}
\begin{split}
L_{sup} &= L_{bce}(\text{Pred}, y) + L_{dice}(\text{Pred}, y) \\
&\quad + L_{bce}(P_{fg}, y_{fg}) + L_{bce}(P_{bg}, y_{bg})
\end{split}
\end{equation}
where $y_{fg} = y$ and $y_{bg} = 1 - y$. The first two terms supervise the main prediction, while the last two terms enforce explicit foreground/background semantics on the disentangled auxiliary branches. The orthogonality loss $L_{orth}$ is defined in Section~\ref{sec:ofdm}. The consistency loss is
\begin{equation}
L_{cons} = \left\| \text{Pred} - P_{Aug} \right\|_2^2
\end{equation}

These losses jointly encourage the model to learn a clear and stable foreground-background representation under labeled supervision.

After Stage I, we introduce unlabeled data $D_U$ for semi-supervised optimization. A difference between the two stages lies in how the perturbed prediction $P_{Aug}$ is supervised. In Stage I, ${Pred}$ is directly constrained by the ground-truth label, and $P_{Aug}$ is optimized through a consistency loss with Pred. In Stage II, ground-truth labels are unavailable for unlabeled samples. We therefore generate pseudo-labels $\hat{y}$ from Pred, retain only high-confidence pixels through confidence filtering, and further modulate the retained regions using the reliability map $M_{rel}$. Accordingly, the supervision on the perturbed branch changes from labeled consistency learning in Stage I to reliability-weighted pseudo-label supervision in Stage II. To gradually increase the contribution of unlabeled supervision, we use a time-dependent ramp-up coefficient:
\begin{equation}
\lambda(t) = \lambda_{\max} \cdot e^{-5(1 - t/t_{\max})^2}
\end{equation}
where $t$ denotes the current training step and $t_{\max}$ denotes the total number of training steps. The total Stage-II objective is
\begin{equation}
L_{stage2} = L_{sup} + \lambda_{orth} L_{orth} + \lambda_{cons} L_{cons} + \lambda(t) L_{unlab}
\end{equation}

The $L_{unlab}$ is defined in Section~\ref{sec:reliability}. This formulation preserves the labeled supervision and regularization terms used in Stage I, while further introducing reliability-weighted unlabeled learning in Stage II. As a result, the model continues to maintain foreground-background disentanglement and perturbation consistency, while gradually exploiting unlabeled data through confidence-filtered and reliability-aware pseudo-label supervision.

\section{Experiments} \label{sec:experiments}

\subsection{Datasets and Evaluation Metrics} \label{sec:datasets_metrics}

\textbf{Datasets.} We evaluate our method on four public medical image segmentation datasets. All experiments are conducted in a slice-wise setting. Among them, ISIC-2016 and Kvasir-SEG are binary segmentation datasets, while Synapse and ACDC are multi-class segmentation datasets.

(1) \textbf{ISIC-2016 \cite{DBLP:journals/corr/GutmanCCHMMH16}:} This dataset contains dermoscopy images and corresponding lesion annotations for skin cancer diagnosis. Its main challenges arise from severe noise interference, such as hair occlusion and imaging artifacts. The dataset includes 900 training images and 379 testing images.

(2) \textbf{Kvasir-SEG \cite{10.1007/978-3-030-37734-2_37}:} This dataset consists of 1,000 polyp images with ground-truth masks. It is challenging because of complex background textures, where the mucosa often shares high visual similarity with polyps, as well as large morphological variation across lesions. We randomly use 80\% of the images for training and the remaining 20\% for testing.

(3) \textbf{Synapse \cite{landman2015miccai}:} The Synapse multi-organ segmentation dataset contains 30 abdominal CT scans with annotations for 13 organs. Following the standard split, we use 20 scans for training and 10 scans for testing.

(4) \textbf{ACDC \cite{bernard2018deep}:} The ACDC dataset is a four-class cardiac segmentation benchmark, including background, right ventricle, left ventricle, and myocardium. It contains scans from 100 patients. Following the standard split, we use 70 patients for training, 10 for validation, and 20 for testing.

\textbf{Evaluation Metrics.} We use different evaluation metrics for binary and multi-class segmentation tasks according to common practice on each benchmark. For the binary segmentation datasets, namely ISIC-2016 and Kvasir-SEG, we report the Dice Similarity Coefficient (DSC), Intersection over Union (IoU), and pixel-level Accuracy (Acc). For multi-class datasets, we report dataset-specific overlap and boundary metrics. Specifically, for Synapse we report Dice and HD95, while for ACDC we report Dice, Jaccard, HD95, and ASD. DSC, IoU, and Jaccard measure the overlap between the predicted mask and the ground truth, while Acc reflects overall pixel-wise classification performance. HD95 and ASD further evaluate boundary accuracy and geometric consistency of the segmentation results.

\subsection{Implementation Details} \label{sec:implementation}

We initialize the SegFormer-B4 backbone with ImageNet-pretrained weights \cite{NEURIPS2021_64f1f27b} to accelerate convergence. We optimize the network using the AdamW optimizer with a weight decay of $1 \times 10^{-4}$. The training process consists of two stages. In Stage I, we train the model for 50 epochs with a learning rate of $1 \times 10^{-4}$. This stage focuses on establishing a stable disentangled foreground-background representation under semantic-frequency perturbations. In Stage II, we continue training for another 50 epochs with a reduced learning rate of $1 \times 10^{-5}$. In this stage, we introduce unlabeled data for semi-supervised optimization, and the unsupervised weight follows a sigmoid ramp-up schedule. For input preprocessing, images from ISIC-2016, Kvasir-SEG, and Synapse are resized to $512 \times 512$, while images from ACDC are resized to $256 \times 256$. The batch size is set to 2. The main hyperparameters are set as $\lambda_{orth} = 0.1$, $\lambda_{cons} = 0.1$, $\lambda_{\max} = 0.1$ and $\tau = 0.95$. We implement all experiments in PyTorch and run them on a single NVIDIA GeForce RTX 4070 Ti Super GPU.

\subsection{Comparisons with State-of-the-Art Methods} \label{sec:comparisons}

\begin{table*}[!t]
\caption{Comparison of Segmentation Performance on ISIC-2016 and Kvasir-SEG Datasets Under Different Labeled Data Ratios (10\% and 20\%). The Best Results Are Highlighted in Bold.}
\label{tab_1}
\centering
\renewcommand{\arraystretch}{1.1} 
\begin{tabular}{l c c c c c c c}
\hline
\multirow{2}{*}{Methods} & \multirow{2}{*}{Ratio} & \multicolumn{3}{c}{ISIC-2016} & \multicolumn{3}{c}{Kvasir-SEG} \\ \cline{3-8} 
 & & DICE(\%)$\uparrow$ & ACC(\%)$\uparrow$ & IOU(\%)$\uparrow$ & DICE(\%)$\uparrow$ & ACC(\%)$\uparrow$ & IOU(\%)$\uparrow$ \\ \hline
SegFormer-B4 \cite{NEURIPS2021_64f1f27b} & 100\% & 92.88 & 95.93 & 87.31 & 91.79 & 97.48 & 86.86 \\ \hline
SegFormer-B4 \cite{NEURIPS2021_64f1f27b} & 10\% & 87.15 & 93.51 & 79.54 & 78.83 & 94.52 & 71.07 \\
MT \cite{tarvainen2017mean} & 10\% & 88.01 & 94.03 & 81.42 & 83.45 & 94.97 & 76.32 \\
SASSNet \cite{10.1007/978-3-030-59710-8_54} & 10\% & 88.42 & 93.78 & 81.46 & 82.78 & 94.78 & 75.21 \\
ST++ \cite{Yang_2022_CVPR} & 10\% & 88.17 & 94.98 & 82.46 & 86.23 & 95.31 & 80.41 \\
URPC \cite{10.1007/978-3-030-87196-3_30} & 10\% & 88.12 & 93.74 & 81.21 & 82.43 & 94.69 & 75.49 \\
CCT \cite{Ouali_2020_CVPR} & 10\% & 88.22 & 93.78 & 81.34 & 83.74 & 94.88 & 76.45 \\
SLCNet \cite{10.1007/978-3-031-16452-1_14} & 10\% & 88.23 & 94.82 & 82.09 & 83.43 & 94.79 & 75.64 \\
DTC \cite{Luo_Chen_Song_Wang_2021} & 10\% & 87.94 & 93.62 & 81.23 & 80.75 & 94.60 & 73.84 \\
DMT \cite{feng2022dmt} & 10\% & 88.03 & 94.93 & 82.34 & 84.98 & 94.89 & 77.23 \\
BCP \cite{Bai_2023_CVPR} & 10\% & 89.02 & 95.09 & 82.93 & 85.79 & 95.09 & 78.43 \\
ARCO \cite{NEURIPS2023_1f7e6d5c} & 10\% & 89.52 & 95.45 & 84.26 & 87.96 & 96.45 & 81.52 \\
DAW \cite{NEURIPS2023_c28ef844} & 10\% & 89.67 & 95.52 & 84.31 & 88.24 & 96.54 & 81.63 \\
FRCNet \cite{10.1007/978-3-031-72111-3_29} & 10\% & 90.77 & 95.41 & 84.70 & 88.77 & 96.49 & 82.61 \\
AdaptFRCNet \cite{HE2025103626} & 10\% & 91.37 & 95.70 & 85.30 & 89.35 & 96.71 & 83.36 \\
\textbf{Ours} & \textbf{10\%} & \textbf{91.86} & \textbf{96.09} & \textbf{85.76} & \textbf{90.19} & \textbf{96.86} & \textbf{84.31} \\ \hline
SegFormer-B4 \cite{NEURIPS2021_64f1f27b} & 20\% & 88.65 & 94.32 & 81.58 & 82.16 & 95.27 & 74.69 \\
MT \cite{tarvainen2017mean} & 20\% & 89.83 & 95.43 & 83.15 & 84.45 & 95.47 & 77.43 \\
SASSNet \cite{10.1007/978-3-030-59710-8_54} & 20\% & 89.94 & 95.19 & 83.67 & 83.97 & 95.56 & 77.21 \\
ST++ \cite{Yang_2022_CVPR} & 20\% & 90.21 & 95.59 & 85.23 & 87.94 & 96.36 & 82.17 \\
URPC \cite{10.1007/978-3-030-87196-3_30} & 20\% & 89.91 & 95.21 & 82.81 & 85.12 & 95.52 & 78.21 \\
CCT \cite{Ouali_2020_CVPR} & 20\% & 89.59 & 95.42 & 82.99 & 84.89 & 95.78 & 77.94 \\
SLCNet \cite{10.1007/978-3-031-16452-1_14} & 20\% & 89.34 & 95.34 & 84.54 & 85.42 & 95.74 & 78.35 \\
DTC \cite{Luo_Chen_Song_Wang_2021} & 20\% & 89.79 & 95.39 & 83.32 & 84.22 & 95.65 & 78.01 \\
DMT \cite{feng2022dmt} & 20\% & 89.21 & 95.56 & 84.32 & 86.47 & 95.83 & 79.32 \\
BCP \cite{Bai_2023_CVPR} & 20\% & 89.98 & 95.32 & 85.01 & 87.43 & 96.23 & 80.45 \\
ARCO \cite{NEURIPS2023_1f7e6d5c} & 20\% & 92.43 & 96.28 & 86.21 & 90.19 & 96.89 & 84.85 \\
DAW \cite{NEURIPS2023_c28ef844} & 20\% & 92.31 & 96.27 & 86.13 & 90.54 & 96.93 & 84.96 \\
FRCNet \cite{10.1007/978-3-031-72111-3_29} & 20\% & 92.44 & 96.30 & 86.71 & 90.62 & 97.22 & 85.01 \\
AdaptFRCNet \cite{HE2025103626} & 20\% & 92.53 & 96.29 & 86.83 & 90.71 & 97.32 & 85.23 \\
\textbf{Ours} & \textbf{20\%} & \textbf{92.62} & \textbf{96.48} & \textbf{86.87} & \textbf{91.16} & \textbf{97.36} & \textbf{85.44} \\ \hline
\end{tabular}
\end{table*}

\begin{figure*}[!t]
\centering
\includegraphics[width=\textwidth]{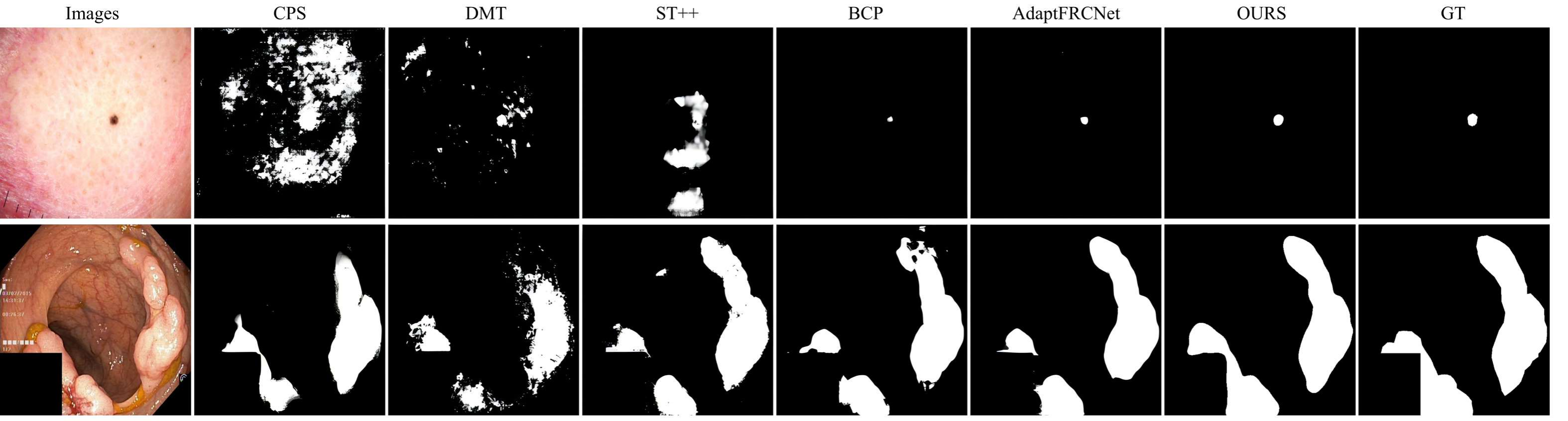}
\caption{Visual results of different methods on skin lesion and polyp segmentation with 10\% labeled data. The first row shows a tiny skin lesion case, and the second row shows a polyp case with a complex shape.}
\label{fig_3}
\end{figure*}

\subsubsection{Comparison on Binary Segmentation Datasets} \label{sec:comp_binary}

Table~\ref{tab_1} reports the quantitative comparison on two binary medical image segmentation datasets, ISIC-2016 and Kvasir-SEG, under 10\% and 20\% labeled-data settings. To ensure a fair comparison and maintain consistency with prior work, the baseline results listed in Table~\ref{tab_1} are quoted from the published AdaptFRCNet \cite{HE2025103626}. As shown in Table~\ref{tab_1}, under both 10\% and 20\% labeled data settings, our method achieves the best performance on both the ISIC-2016 and Kvasir-SEG datasets.

On the ISIC-2016 dataset, with only 10\% labeled data, our method achieves the Dice score of 91.86\%, the Acc score of 96.09\%, and the IoU score of 85.76\%, outperforming AdaptFRCNet by 0.49, 0.39, and 0.46 percentage points in Dice, Acc, and IoU, respectively. When the proportion of labeled data increases to 20\%, our method achieves the Dice score of 92.62\%, the Acc score of 96.48\%, and the IoU score of 86.87\%, which are 0.09, 0.19, and 0.04 percentage points higher than those of AdaptFRCNet, respectively. On the Kvasir-SEG dataset, with 10\% labeled data, our method achieves the Dice score of 90.19\%, the Acc score of 96.86\%, and the IoU score of 84.31\%, surpassing AdaptFRCNet by 0.84, 0.15, and 0.95 percentage points in Dice, Acc, and IoU, respectively. With 20\% labeled data, our method further achieves the Dice score of 91.16\%, the Acc score of 97.36\%, and the IoU score of 85.44\%, exceeding AdaptFRCNet by 0.45, 0.04, and 0.21 percentage points, respectively.

These results demonstrate that our method can effectively leverage a large amount of unlabeled data and substantially reduce the dependence on labeled data. Even with only 20\% labeled data, our results are already very close to those obtained under the fully supervised setting.

For qualitative analysis, Fig.~\ref{fig_3} presents visual comparisons of two challenging binary segmentation cases. In the first row, which corresponds to a tiny skin lesion, several competing methods produce excessive false positives or fail to completely segment the lesion, whereas our method generates a compact prediction that closely matches the ground truth. In the second row, which corresponds to a polyp case with a complex shape, some methods miss parts of the target region or produce incomplete segmentation masks with blurry boundaries. In contrast, our method yields a more complete segmentation result with clearer boundaries and better consistency with the ground truth. These qualitative results are consistent with the quantitative results in Table~\ref{tab_1} and further demonstrate the effectiveness of our method in handling challenging binary segmentation cases.

\begin{table}[!t]
\caption{Quantitative comparison on the Synapse multi-organ CT dataset under 10\% and 20\% labeled settings. The best results are highlighted in bold.}
\label{tab_2}
\centering
\renewcommand{\arraystretch}{1.1} 
\begin{tabular}{l c c c}
\hline
Methods & Ratio & DICE(\%)$\uparrow$ & HD95(mm)$\downarrow$ \\ \hline
SegFormer-B4 \cite{NEURIPS2021_64f1f27b} & 100\% & 65.37 & 22.15 \\ \hline
SegFormer-B4 \cite{NEURIPS2021_64f1f27b} & 10\% & 43.51 & 33.23 \\
MT \cite{tarvainen2017mean} & 10\% & 56.23 & 29.26 \\
SASSNet \cite{10.1007/978-3-030-59710-8_54} & 10\% & 57.42 & 29.11 \\
CCT \cite{Ouali_2020_CVPR} & 10\% & 57.89 & 28.42 \\
URPC \cite{10.1007/978-3-030-87196-3_30} & 10\% & 58.41 & 28.32 \\
DTC \cite{Luo_Chen_Song_Wang_2021} & 10\% & 58.79 & 28.16 \\
CPS \cite{chen2021semi} & 10\% & 57.39 & 27.89 \\
SLCNet \cite{10.1007/978-3-031-16452-1_14} & 10\% & 58.13 & 26.98 \\
DMT \cite{feng2022dmt} & 10\% & 58.46 & 26.85 \\
ST++ \cite{Yang_2022_CVPR} & 10\% & 57.64 & 26.75 \\
BCP \cite{Bai_2023_CVPR} & 10\% & 58.96 & 26.15 \\
ARCO \cite{NEURIPS2023_1f7e6d5c} & 10\% & 59.11 & 25.98 \\
DAW \cite{NEURIPS2023_c28ef844} & 10\% & 59.32 & 25.83 \\
FRCNet \cite{10.1007/978-3-031-72111-3_29} & 10\% & 60.11 & 25.66 \\
AdaptFRCNet \cite{HE2025103626} & 10\% & 61.34 & 25.11 \\
\textbf{Ours} & \textbf{10\%} & \textbf{62.35} & \textbf{24.25} \\ \hline
SegFormer-B4 \cite{NEURIPS2021_64f1f27b} & 20\% & 45.75 & 31.65 \\
MT \cite{tarvainen2017mean} & 20\% & 57.89 & 28.56 \\
SASSNet \cite{10.1007/978-3-030-59710-8_54} & 20\% & 58.36 & 28.45 \\
CCT \cite{Ouali_2020_CVPR} & 20\% & 58.74 & 27.69 \\
URPC \cite{10.1007/978-3-030-87196-3_30} & 20\% & 59.02 & 27.68 \\
DTC \cite{Luo_Chen_Song_Wang_2021} & 20\% & 58.95 & 27.45 \\
SLCNet \cite{10.1007/978-3-031-16452-1_14} & 20\% & 59.03 & 25.98 \\
DMT \cite{feng2022dmt} & 20\% & 59.11 & 25.85 \\
ST++ \cite{Yang_2022_CVPR} & 20\% & 58.53 & 26.24 \\
BCP \cite{Bai_2023_CVPR} & 20\% & 59.44 & 25.73 \\
ARCO \cite{NEURIPS2023_1f7e6d5c} & 20\% & 59.95 & 25.56 \\
DAW \cite{NEURIPS2023_c28ef844} & 20\% & 60.42 & 25.32 \\
FRCNet \cite{10.1007/978-3-031-72111-3_29} & 20\% & 60.24 & 26.01 \\
AdaptFRCNet \cite{HE2025103626} & 20\% & 62.26 & 24.16 \\
\textbf{Ours} & \textbf{20\%} & \textbf{63.29} & \textbf{23.10} \\ \hline
\end{tabular}
\end{table}

\begin{figure*}[!t]
\centering
\includegraphics[width=\textwidth]{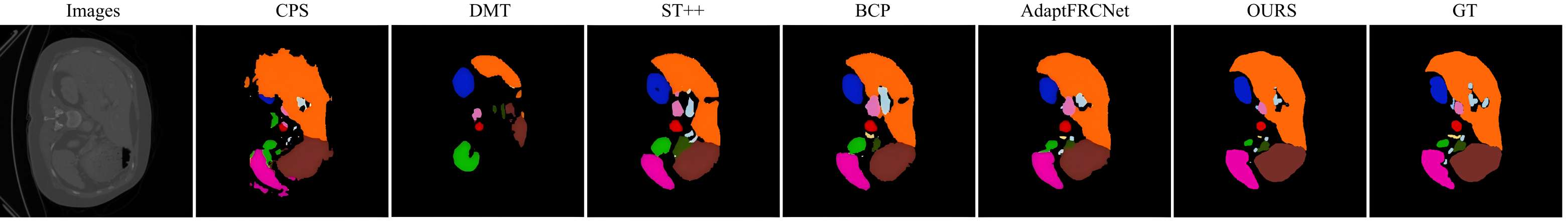}
\caption{Visual comparison of multi-organ segmentation results on the Synapse dataset.}
\label{fig_4}
\end{figure*}

\subsubsection{Comparison on Multi-class Segmentation Datasets} \label{sec:comp_multi}

The quantitative results on the multi-class segmentation datasets are reported in Table~\ref{tab_2} and Table~\ref{tab_3}. For consistency with prior work, the baseline results in Table~\ref{tab_2} are quoted from the published AdaptFRCNet \cite{HE2025103626}, while the baseline results in Table~\ref{tab_3} are quoted from the published BCP \cite{Bai_2023_CVPR} and the KnowSAM result is quoted from its original paper \cite{huang2025learnable}. It is worth noting that, in Table~\ref{tab_3}, the rows "U-Net (5\%)" and "U-Net (10\%)" denote supervised training using only 5\% and 10\% labeled data, respectively. In contrast, the other methods are semi-supervised and are trained using the corresponding labeled subset together with the remaining unlabeled data.

On the Synapse multi-organ CT dataset, with 10\% labeled data, our method achieves the Dice score of 62.35\% and the HD95 of 24.25 mm, outperforming AdaptFRCNet by 1.01 percentage points in Dice and reducing HD95 by 0.86 mm. With 20\% labeled data, our method further achieves the Dice score of 63.29\% and the HD95 of 23.10 mm, which are 1.03 percentage points higher and 1.06 mm lower than those of AdaptFRCNet, respectively. These results show that our method consistently achieves the best performance on the Synapse dataset under both labeled settings.

Fig.~\ref{fig_4} presents a visual comparison of multi-organ segmentation results on the Synapse dataset. It can be observed that several competing methods either miss small organ regions and produce incomplete predictions, or generate inaccurate boundaries in challenging anatomical areas. In contrast, our method produces more complete segmentation results that are more consistent with the ground truth. It also better preserves the structural integrity of multiple organs while reducing confusion among adjacent anatomical regions.

\begin{table}[!t]
\caption{Quantitative comparison on the ACDC dataset under 5\% and 10\% labeled settings. The rows "U-Net (5\%)" and "U-Net (10\%)" denote supervised training using only the corresponding labeled subsets, while the remaining methods are semi-supervised and additionally use the remaining unlabeled data. Our method is evaluated with both U-Net and SegFormer-B4 backbones. The best results are highlighted in bold.}
\label{tab_3}
\centering
\renewcommand{\arraystretch}{1.1} 
\resizebox{\columnwidth}{!}{
\begin{tabular}{l c c c c c}
\hline
Methods & Ratio & DICE(\%)$\uparrow$ & Jaccard(\%)$\uparrow$ & HD95(mm)$\downarrow$ & ASD(mm)$\downarrow$ \\ \hline
U-Net \cite{ronneberger2015u} & 100\% & 91.44 & 84.59 & 4.30 & 0.99 \\ \hline
U-Net \cite{ronneberger2015u} & 5\% & 47.83 & 37.01 & 31.16 & 12.62 \\
UA-MT \cite{10.1007/978-3-030-32245-8_67} & 5\% & 46.04 & 35.97 & 20.08 & 7.75 \\
SASSNet \cite{10.1007/978-3-030-59710-8_54} & 5\% & 57.77 & 46.14 & 20.05 & 6.06 \\
DTC \cite{Luo_Chen_Song_Wang_2021} & 5\% & 56.90 & 45.67 & 23.36 & 7.39 \\
URPC \cite{10.1007/978-3-030-87196-3_30} & 5\% & 55.87 & 44.64 & 13.60 & 3.74 \\
MC-Net \cite{wu2021semi} & 5\% & 62.85 & 52.29 & 7.62 & 2.33 \\
SS-Net \cite{wu2022exploring} & 5\% & 65.83 & 55.38 & 6.67 & 2.28 \\
BCP \cite{Bai_2023_CVPR} & 5\% & 87.59 & 78.67 & \textbf{1.90} & \textbf{0.67} \\
\textbf{Ours (U-Net)} & \textbf{5\%} & \textbf{88.92} & \textbf{83.03} & 2.44 & 1.37 \\
\textbf{Ours (SegFormer-B4)} & \textbf{5\%} & \textbf{89.47} & \textbf{84.76} & 2.18 & 1.21 \\ \hline
U-Net \cite{ronneberger2015u} & 10\% & 79.41 & 68.11 & 9.35 & 2.70 \\
UA-MT \cite{10.1007/978-3-030-32245-8_67} & 10\% & 81.65 & 70.64 & 6.88 & 2.02 \\
SASSNet \cite{10.1007/978-3-030-59710-8_54} & 10\% & 84.50 & 74.34 & 5.42 & 1.86 \\
DTC \cite{Luo_Chen_Song_Wang_2021} & 10\% & 84.29 & 73.92 & 12.81 & 4.01 \\
URPC \cite{10.1007/978-3-030-87196-3_30} & 10\% & 83.10 & 72.41 & 4.84 & 1.53 \\
MC-Net \cite{wu2021semi} & 10\% & 86.44 & 77.04 & 5.50 & 1.84 \\
SS-Net \cite{wu2022exploring} & 10\% & 86.78 & 77.67 & 6.07 & 1.40 \\
BCP \cite{Bai_2023_CVPR} & 10\% & 88.84 & 80.62 & 3.98 & 1.17 \\
KnowSAM \cite{huang2025learnable} & 10\% & 89.56 & 81.66 & 1.28 & 0.36 \\
\textbf{Ours (U-Net)} & \textbf{10\%} & \textbf{89.97} & \textbf{84.68} & 1.79 & 1.02 \\
\textbf{Ours (SegFormer-B4)} & \textbf{10\%} & \textbf{90.42} & \textbf{86.25} & 1.36 & 0.78 \\ \hline
\end{tabular}
}
\end{table}

On the ACDC dataset, we further evaluate our method with a different backbone, namely U-Net, in addition to SegFormer-B4. Under the 5\% labeled setting, Ours (U-Net) achieves a Dice score of 88.92\%, a Jaccard score of 83.03\%, an HD95 of 2.44 mm, and an ASD of 1.37 mm, while Ours (SegFormer-B4) further improves the performance to a Dice score of 89.47\%, a Jaccard score of 84.76\%, an HD95 of 2.18 mm, and an ASD of 1.21 mm. Under the 10\% labeled setting, Ours (U-Net) achieves a Dice score of 89.97\%, a Jaccard score of 84.68\%, an HD95 of 1.79 mm, and an ASD of 1.02 mm, while Ours (SegFormer-B4) further achieves a Dice score of 90.42\%, a Jaccard score of 86.25\%, an HD95 of 1.36 mm, and an ASD of 0.78 mm. Compared with BCP, Ours (SegFormer-B4) improves Dice by 1.88 percentage points and Jaccard by 6.09 percentage points under the 5\% labeled setting, and improves Dice by 1.58 percentage points and Jaccard by 5.63 percentage points while reducing HD95 by 2.62 mm and ASD by 0.39 mm under the 10\% labeled setting. Compared with KnowSAM under the 10\% labeled setting, Ours (SegFormer-B4) further improves the Dice score by 0.86 percentage points and the Jaccard score by 4.59 percentage points. These results show that our method achieves the strongest overlap performance on ACDC under both backbone settings, while maintaining competitive boundary accuracy, demonstrating that its effectiveness is not limited to a specific architecture.

\begin{table}[!t]
\caption{Ablation study of different components on the ISIC-2016 dataset under the 20\% labeled setting. Baseline denotes SegFormer-B4 trained using labeled data only. OFDM denotes the Orthogonal Feature Disentanglement Module. FDM denotes the Feature Disentanglement Module without the orthogonal loss $L_{orth}$. DHF denotes Dual-Head Fusion. DP denotes Semantic-Frequency Dual Perturbation. $M_{rel}$ denotes the reliability assessment mechanism. The best results are highlighted in bold.}
\label{tab_4}
\centering
\renewcommand{\arraystretch}{1.1} 
\scalebox{0.8}{ 
\begin{tabular}{c c c c c c c c}
\hline
\multicolumn{5}{c}{Components} & \multicolumn{3}{c}{Metrics} \\
Baseline & OFDM & DHF & DP & $M_{rel}$ & Dice(\%) & ACC(\%) & IoU(\%) \\ \hline
\checkmark & & & & & 88.65 & 94.32 & 81.58 \\
\checkmark & \checkmark & & & & 91.45 & 94.83 & 85.65 \\
\checkmark & \checkmark & & & \checkmark & 91.89 & 95.10 & 86.07 \\
\checkmark & \checkmark & \checkmark & & \checkmark & 92.43 & 96.32 & 86.73 \\
\checkmark & \checkmark & & \checkmark & \checkmark & 92.40 & 95.34 & 86.59 \\
\checkmark & \checkmark & \checkmark & \checkmark & & 91.85 & 95.19 & 86.03 \\
\checkmark & FDM & \checkmark & \checkmark & \checkmark & 92.31 & 96.28 & 86.43 \\
\checkmark & \checkmark & \checkmark & \checkmark & \checkmark & \textbf{92.62} & \textbf{96.48} & \textbf{86.87} \\ \hline
\end{tabular}
}
\end{table}

\subsection{Ablation Study} \label{sec:ablation}

We conduct a comprehensive ablation study on the ISIC-2016 dataset under the 20\% labeled setting to evaluate the contribution of each core component in OFD-Net. As shown in Table~\ref{tab_4}, the supervised SegFormer-B4 baseline achieves the Dice score of 88.65\%, the Acc score of 94.32\%, and the IoU score of 81.58\%. After introducing OFDM, the model improves by 2.80 percentage points, 0.51 percentage points, and 4.07 percentage points in Dice, Acc, and IoU, respectively, corresponding to relative improvements of 3.16\%, 0.54\%, and 4.99\%. Further adding the reliability assessment mechanism $M_{rel}$ brings additional gains of 0.44 percentage points, 0.27 percentage points, and 0.42 percentage points, respectively, corresponding to relative improvements of 0.48\%, 0.28\%, and 0.49\%. On top of this, incorporating DHF further improves the performance by 0.54 percentage points in Dice, 1.22 percentage points in Acc, and 0.66 percentage points in IoU, corresponding to relative improvements of 0.59\%, 1.28\%, and 0.77\%, respectively. Finally, after introducing DP, the complete model achieves the best overall performance, reaching the Dice score of 92.62\%, the Acc score of 96.48\%, and the IoU score of 86.87\%. Compared with the previous variant, this corresponds to further gains of 0.19 percentage points, 0.16 percentage points, and 0.14 percentage points in Dice, Acc, and IoU, respectively, equivalent to relative improvements of 0.21\%, 0.17\%, and 0.16\%. These results indicate that OFDM, $M_{rel}$, DHF, and DP all contribute positively to the final model, and their combination yields the best overall results.

\begin{figure}[!t]
\centering
\includegraphics[width=\columnwidth]{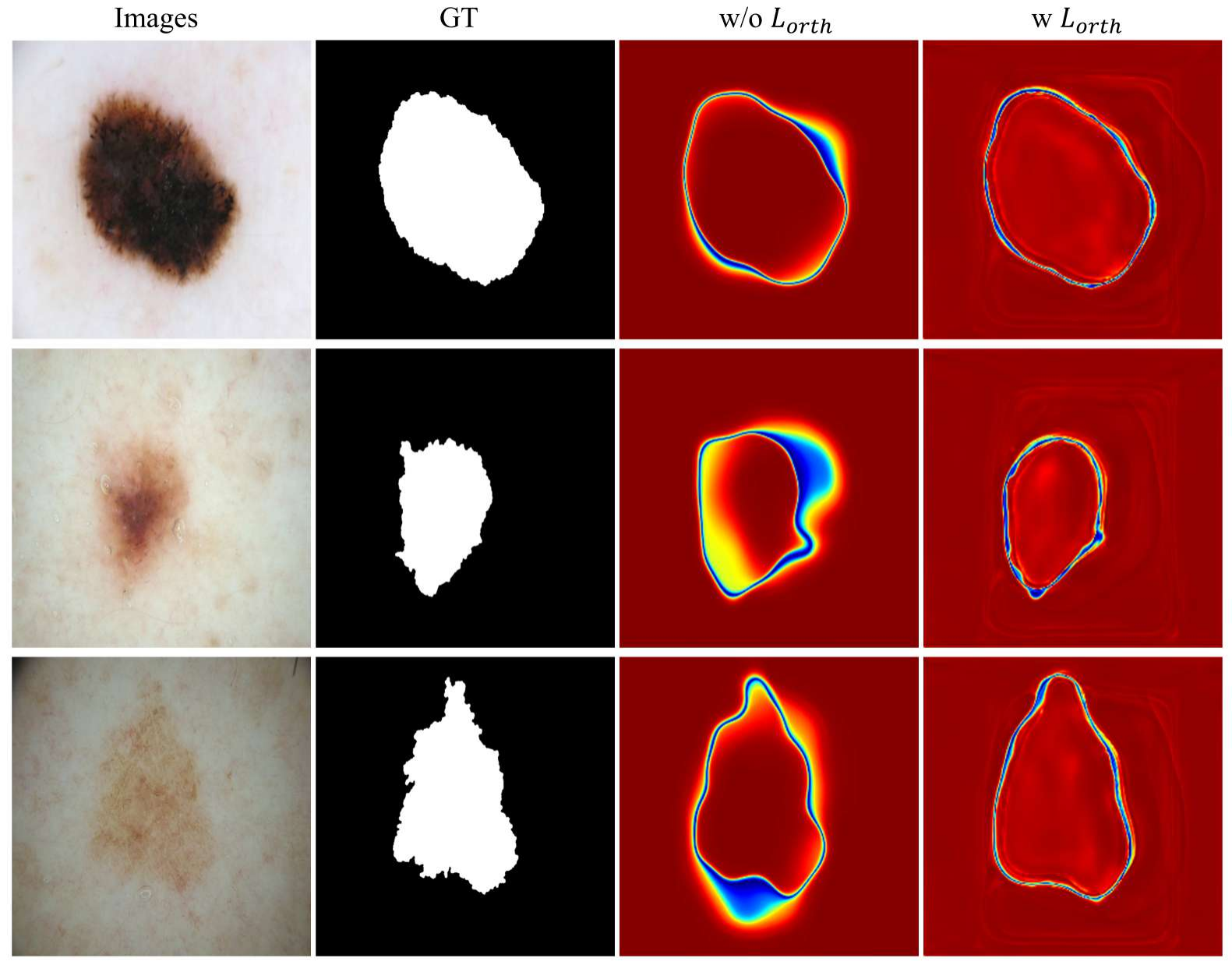}
\caption{Visualization of reliability maps with and without the orthogonal constraint on the ISIC-2016 dataset.}
\label{fig_5}
\end{figure}

\textbf{Effectiveness of OFDM and the orthogonal constraint $L_{orth}$.} Incorporating OFDM into the baseline model improves the segmentation performance, which verifies the importance of explicitly disentangling foreground and background representations. More importantly, the orthogonal constraint is not merely an auxiliary regularization term, but a necessary condition for achieving clean feature separation. When OFDM is replaced by its disentanglement variant without the orthogonal loss $L_{orth}$, the performance drops from the Acc score of 96.48\%, the Dice score of 92.62\%, and the IoU score of 86.87\% to the Acc score of 96.28\%, the Dice score of 92.31\%, and the IoU score of 86.43\%, corresponding to decreases of 0.20, 0.31, and 0.44 percentage points, respectively. This indicates that, without the mutual-exclusivity constraint, foreground and background features remain more entangled at the bottleneck, which weakens the structural guidance provided to the decoder.

The reliability-map visualizations further support this observation. As shown in Fig.~\ref{fig_5}, the variant without the orthogonal constraint produces more diffuse responses, with evident activation leakage outside the lesion region and less precise boundary localization. In contrast, the full model with the orthogonal constraint generates more compact activations that are better aligned with the lesion contour and more consistent with the ground truth. This qualitative comparison suggests that the orthogonal constraint helps suppress redundant background responses and enables a cleaner and more boundary-aware foreground-background separation.

\begin{table}[!t]
\caption{Performance improvements after introducing $M_{rel}$ across different sample-difficulty quartiles on the ISIC-2016 dataset. The quartiles are obtained by ranking all test samples according to the Dice scores of the variant without $M_{rel}$, from low to high. Thus, the bottom 25\% correspond to the hardest samples, while the top 25\% correspond to the easiest samples.}
\label{tab_5}
\centering
\renewcommand{\arraystretch}{1.1} 
\begin{tabular}{l c c c c}
\hline
Metric & Hardest (bottom 25\%) & 25\%-50\% & 50\%-75\% & top 25\% \\ \hline
$\Delta$Dice(\%) & +1.57 & +0.93 & +0.41 & +0.18 \\
$\Delta$IoU(\%) & +1.63 & +1.01 & +0.53 & +0.21 \\ \hline
\end{tabular}
\end{table}

\begin{figure}[!t]
\centering
\includegraphics[width=\columnwidth]{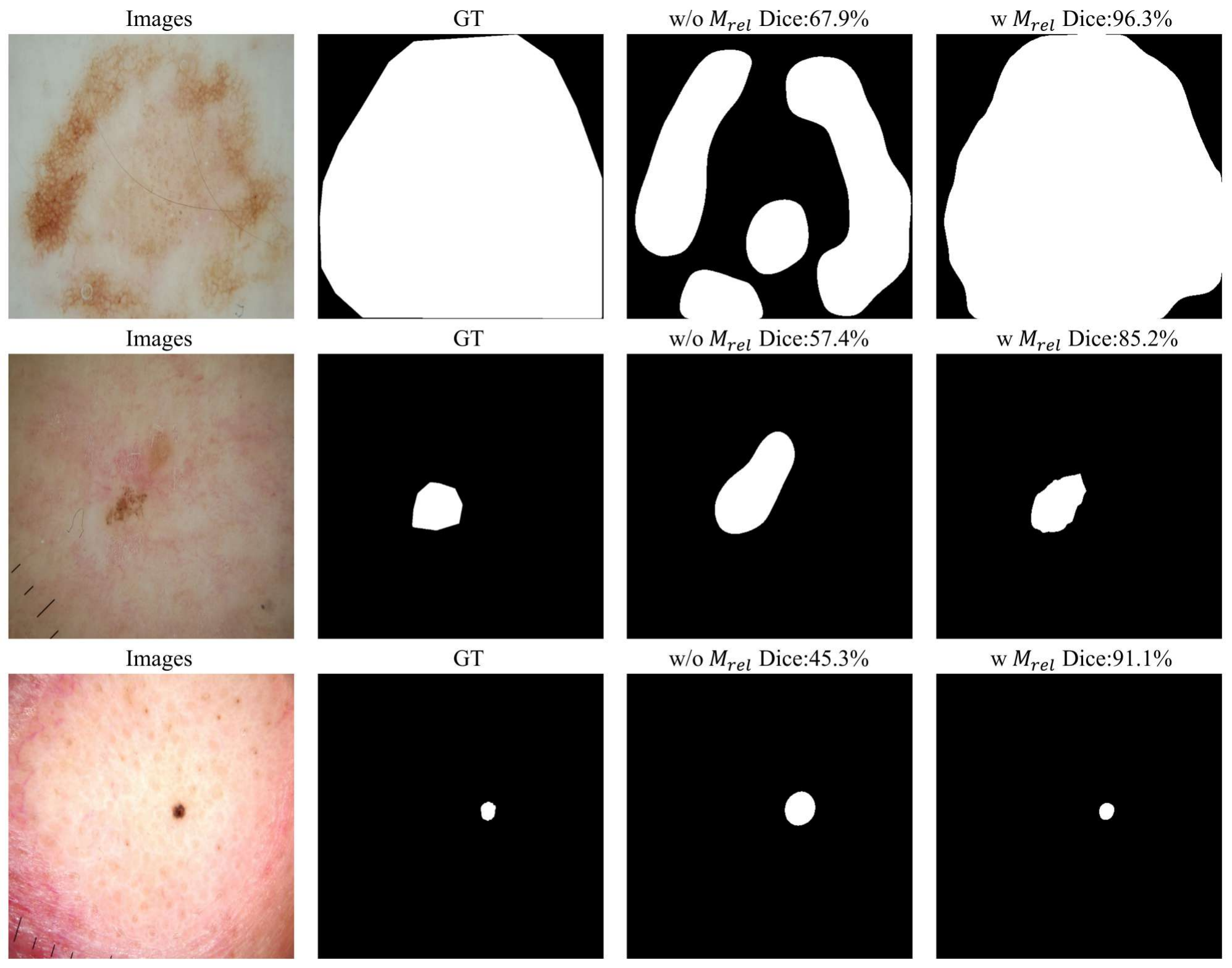}
\caption{Performance improvements after introducing $M_{rel}$ across different sample-difficulty quartiles on the ISIC-2016 dataset.}
\label{fig_6}
\end{figure}

\textbf{Effectiveness of the reliability assessment mechanism.} The disentanglement-based reliability assessment mechanism further improves the effectiveness of pseudo-label learning. As shown in Table~\ref{tab_4}, removing $M_{rel}$ from the full model reduces the Acc score from 96.48\% to 95.19\%, the Dice score from 92.62\% to 91.85\%, and the IoU score from 86.87\% to 86.03\%, corresponding to drops of 1.29, 0.77, and 0.84 percentage points, respectively. This result indicates that relying solely on pseudo-label confidence is insufficient, whereas structural verification based on disentangled foreground-background responses provides a more reliable criterion for unlabeled supervision.

To further investigate the source of this performance gain, we analyze the effect of $M_{rel}$ across different sample-difficulty quartiles on the ISIC-2016 dataset. Specifically, we rank all test samples according to the Dice scores of the variant without $M_{rel}$ from low to high, where the bottom 25\% therefore correspond to the most difficult samples. As shown in Table~\ref{tab_5}, the improvement brought by $M_{rel}$ increases monotonically as the sample difficulty increases. Specifically, the Dice gain rises from +0.18\% on the easiest quartile to +0.41\%, +0.93\%, and finally +1.57\% on the hardest quartile, while the corresponding IoU gain increases from +0.21\% to +0.53\%, +1.01\%, and +1.63\%.

The qualitative results further support this observation. As shown in Fig.~\ref{fig_6}, without $M_{rel}$, the model tends to produce fragmented predictions, inaccurate object scales, or obvious false positives on hard cases. After introducing $M_{rel}$ these failure cases are substantially corrected, and the predictions become more complete and more consistent with the ground truth. Therefore, the primary role of $M_{rel}$ is not merely to provide a uniform average improvement, but to suppress catastrophic failures caused by unreliable pseudo-labels on difficult samples.

\begin{table}[!t]
\caption{Ablation study of semantic and frequency perturbations on the ISIC-2016 dataset under the 20\% labeled setting. ``w/o Semantic'' denotes removing the semantic perturbation only, ``w/o Frequency'' denotes removing the frequency perturbation only, and ``w/o DP'' denotes removing both perturbations. The best results are highlighted in bold.}
\label{tab_6}
\centering
\renewcommand{\arraystretch}{1.1} 
\begin{tabular}{c c c c}
\hline
Model & Dice(\%) & Acc(\%) & IoU(\%) \\ \hline
w/o DP & 92.40 & 95.34 & 86.59 \\
w/o Semantic & 92.53 & 96.42 & 86.78 \\
w/o Frequency & 92.49 & 96.38 & 86.75 \\
OFD-Net & \textbf{92.62} & \textbf{96.48} & \textbf{86.87} \\ \hline
\end{tabular}
\end{table}

\begin{figure}[!t]
\centering
\includegraphics[width=\columnwidth]{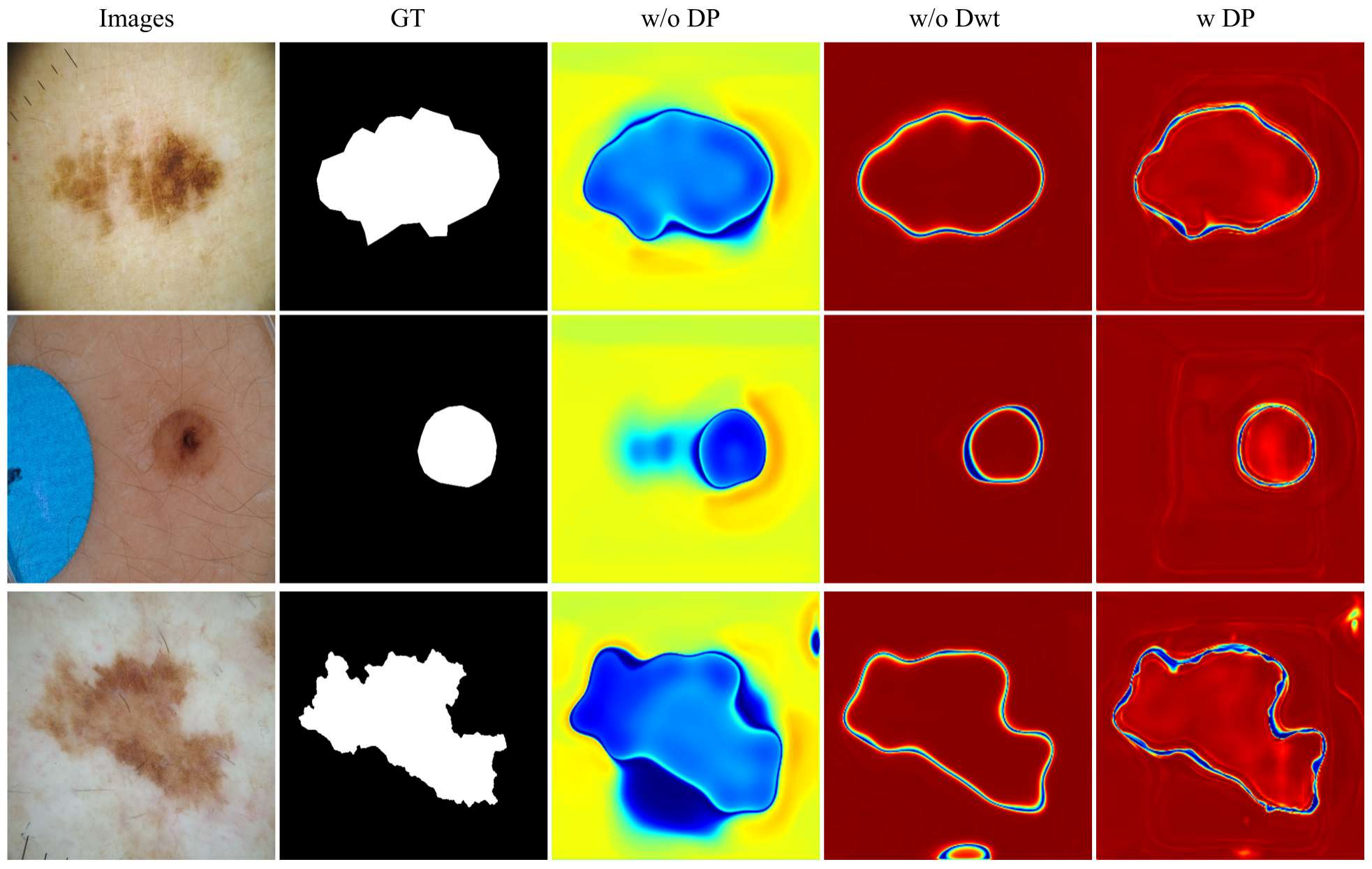}
\caption{Visualization of response maps under different perturbation settings on the ISIC-2016 dataset.}
\label{fig_7}
\end{figure}

\textbf{Effectiveness of dual perturbation.} The dual-perturbation strategy further improves the robustness of the proposed model. As shown in Table~\ref{tab_6}, compared with the full model, removing the semantic perturbation decreases the Dice score from 92.62\% to 92.53\%, the Acc score from 96.48\% to 96.42\%, and the IoU score from 86.87\% to 86.78\%, corresponding to drops of 0.09, 0.06, and 0.09 percentage points, respectively. Removing the frequency perturbation decreases the Dice score to 92.49\%, the Acc score to 96.38\%, and the IoU score to 86.75\%, corresponding to drops of 0.13, 0.10, and 0.12 percentage points, respectively. When both perturbations are removed, the performance further declines to the Dice score of 92.40\%, the Acc score of 95.34\%, and the IoU score of 86.59\%, corresponding to larger drops of 0.22, 1.14, and 0.28 percentage points, respectively. These results show that both perturbations contribute positively to the final performance, and removing both causes a larger degradation than removing either one alone.

As shown in Fig.~\ref{fig_7}, removing both perturbations leads to more diffuse and less localized responses, while removing only the frequency perturbation still weakens the boundary-related activation around the lesion contour. In contrast, the full model with dual perturbation produces sharper and more compact responses that are better aligned with the lesion boundary. This visual comparison further suggests that the two perturbations are complementary in improving feature robustness.

\textbf{Effectiveness of dual-head fusion.} After introducing OFDM and $M_{rel}$, adding the dual-head fusion module further improves the Acc score from 95.10\% to 96.32\%, the Dice score from 91.89\% to 92.43\%, and the IoU score from 86.07\% to 86.73\%, corresponding to gains of 1.22, 0.54, and 0.66 percentage points, respectively. This result indicates that fusing the decoder output with the disentanglement branch is beneficial, as it combines the detailed spatial information provided by the decoder with the clearer structural prior extracted by the disentanglement head, thereby producing more accurate segmentation boundaries.

\begin{figure}[!t]
\centering
\includegraphics[width=\columnwidth]{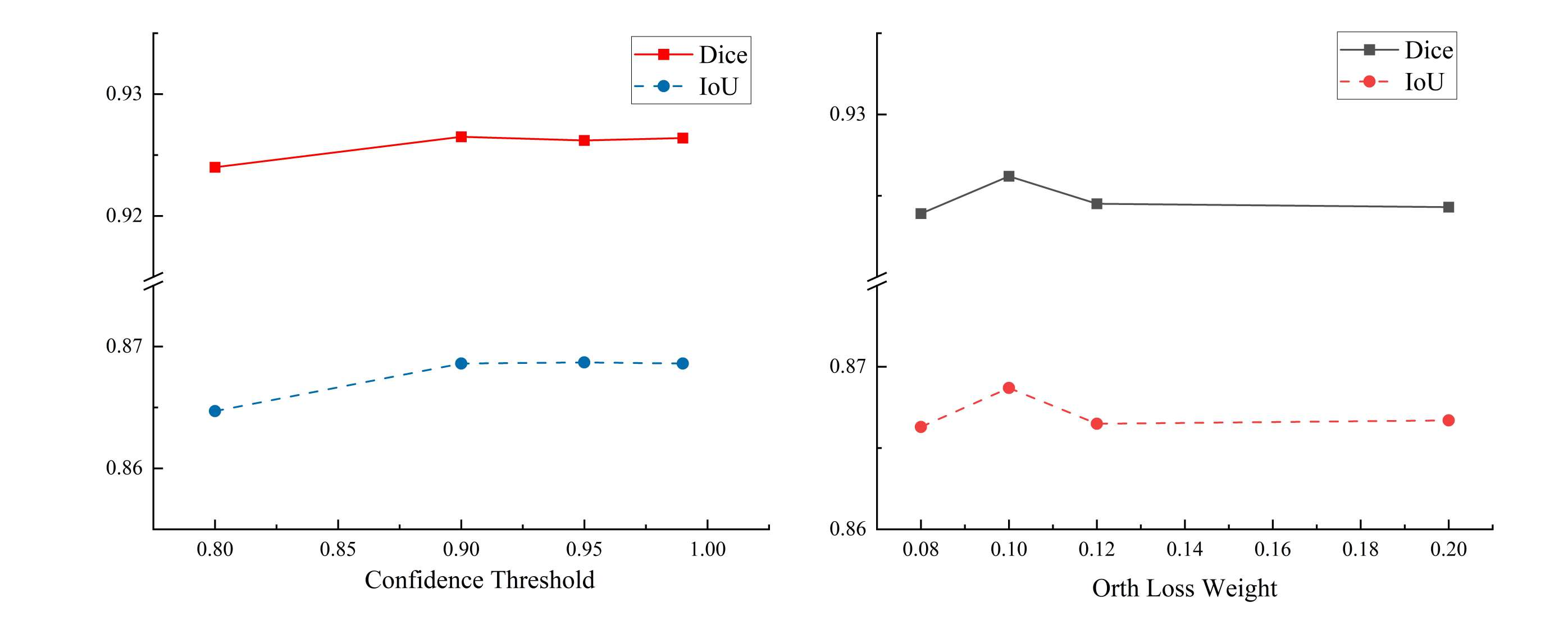}
\caption{Sensitivity analysis of the pseudo-label confidence threshold and the orthogonal loss weight on the ISIC-2016 dataset.}
\label{fig_8}
\end{figure}

\textbf{Hyperparameter robustness.} We further examine the sensitivity of the main hyperparameters, including the pseudo-label confidence threshold $\tau$ and the orthogonal loss weight $\lambda_{orth}$. As shown in Fig.~\ref{fig_8}, the model performance remains highly stable over a range of tested values. When the confidence threshold varies from 0.80 to 0.99, both the Dice score and the IoU score exhibit only minor fluctuations. A similar trend can be observed when the orthogonal loss weight changes from 0.08 to 0.20. In both cases, the maximum deviation in Dice remains within 0.2 percentage points. These results indicate that the effectiveness of OFDM, the orthogonal constraint, and the reliability mechanism does not rely on delicate hyperparameter tuning, which further supports the robustness of the proposed design in practical applications.

\begin{table}[!t]
\caption{Backbone ablation of OFD-Net on the ISIC-2016 skin lesion dataset.}
\label{tab_7}
\centering
\renewcommand{\arraystretch}{1.1} 
\begin{tabular}{l c c c}
\hline
Backbone & Dice(\%) & ACC(\%) & IoU(\%) \\ \hline
U-Net & 91.52 & 95.14 & 81.96 \\
SegFormer-B4 & \textbf{92.62} & \textbf{96.48} & \textbf{86.87} \\ \hline
\end{tabular}
\end{table}

\textbf{Backbone ablation.} We further conduct a backbone ablation on the ISIC-2016 skin lesion dataset to examine whether the effectiveness of OFD-Net depends on a specific encoder architecture. As shown in Table~\ref{tab_7}, when using U-Net as the backbone, OFD-Net achieves the Dice score of 91.52\%, the Acc score of 95.14\%, and the IoU score of 81.96\%. Notably, even with the simpler U-Net backbone, OFD-Net still achieves a Dice score that is higher than that of many compared semi-supervised methods in Table~\ref{tab_1}, such as BCP, which reports a Dice score of 89.98\% under the same 20\% labeled setting. Replacing U-Net with SegFormer-B4 further improves the performance to the Dice score of 92.62\%, the Acc score of 96.48\%, and the IoU score of 86.87\%, corresponding to gains of 1.10, 1.34, and 4.91 percentage points, respectively. These results indicate that OFD-Net is not limited to a specific backbone architecture, while SegFormer-B4 provides the strongest overall performance in our setting.

\section{Discussion} \label{sec:discussion}

Unlike conventional Mean Teacher (MT)-style semi-supervised segmentation frameworks, which obtain unlabeled supervision by enforcing consistency between a student network and an EMA-updated teacher network, OFD-Net does not rely on an external teacher branch. Instead, it establishes unlabeled supervision within a single network by explicitly learning disentangled foreground-background representations and using them to assess pseudo-label reliability. In this sense, the essential difference from traditional MT-style methods is not merely architectural simplification, but the replacement of externally provided consistency supervision with internally learned, structure-aware reliability modeling.

The overall performance gain of OFD-Net arises from the coordinated effect of foreground-background disentanglement, reliability-aware pseudo-label learning, and semantic-frequency dual perturbation. Rather than acting as isolated add-on modules, these components form a coherent learning pipeline in which disentanglement provides the structural basis, decoder guidance propagates this prior to downstream prediction, and reliability modeling verifies unlabeled supervision based on the learned structure. This interpretation is consistent with the ablation results, where OFDM, the orthogonal constraint, the reliability mechanism, dual-head fusion, and dual perturbation all contribute positively to the final performance.

$M_{rel}$ serves two purposes in OFD-Net: it improves the overall segmentation performance and, more importantly, it suppresses severe pseudo-label errors in structurally ambiguous or failure-prone regions. This interpretation is supported by both the quartile analysis and the qualitative comparisons, which show that the improvement introduced by $M_{rel}$ becomes more evident as sample difficulty increases. Such a trend is consistent with the original design motivation of the proposed reliability mechanism. When pseudo-label reliability cannot be reliably judged by confidence alone, the additional structure-aware criterion provided by $M_{rel}$ can more effectively filter unreliable supervision.

The orthogonal constraint also plays a more substantial role than a conventional auxiliary regularizer. Its effect is not limited to improving the separability of bottleneck features. More importantly, it provides a cleaner foreground-background structural basis for both decoder guidance and reliability estimation. Once this constraint is removed, the responses become more diffuse and less aligned with object boundaries, which weakens structural discrimination and subsequently reduces the quality of pseudo-label verification. This result suggests that the effectiveness of the reliability mechanism depends on the cleanness of the learned disentangled structure itself.

The dual perturbation strategy is introduced because the proposed framework must remain stable under different sources of variation during both labeled structural learning and unlabeled optimization. Semantic perturbation mainly addresses appearance-level style shifts, such as changes in color, intensity, or contrast, which may otherwise cause the model to overfit superficial visual statistics and weaken the transfer of reliable structural cues from labeled to unlabeled data. Frequency perturbation addresses a different problem: medical images often contain redundant texture responses, noise, and fine-scale interference that do not necessarily reflect the true target boundary. By perturbing the image in the frequency domain, the model is encouraged to reduce its dependence on unstable texture patterns and rely more on structurally meaningful cues. Therefore, the two perturbations are complementary rather than redundant: one improves robustness to appearance variation, while the other improves robustness to frequency-domain interference. Their combination leads to more stable representation learning than either perturbation alone, which is consistent with the ablation and visualization results.

Despite these advantages, the current study still has several limitations. First, the effectiveness of the proposed reliability mechanism depends on the quality of the learned foreground-background disentanglement; if the bottleneck features are not sufficiently separated, the reliability map may become less discriminative. Second, the present study mainly focuses on 2D slice-wise segmentation, and the extension to fully 3D semi-supervised medical image segmentation remains to be explored. Finally, although OFD-Net avoids maintaining an additional EMA teacher, it still introduces extra modules for disentanglement, guidance, and reliability estimation, which increases implementation complexity. Future work will investigate more lightweight reliability modeling, broader validation across different imaging modalities and anatomical structures, and a more unified extension to volumetric and large-scale multi-class segmentation settings.

\section{Conclusion} \label{sec:conclusion}

In this paper, we proposed OFD-Net, a teacher-free single-network framework for semi-supervised medical image segmentation. Different from conventional MT-style methods that rely on teacher-student consistency to provide unlabeled supervision, OFD-Net constructs semi-supervised learning within a single network by integrating orthogonal foreground-background disentanglement, disentanglement-guided decoding, reliability-aware pseudo-label weighting, and semantic-frequency dual perturbation. In this way, externally provided teacher supervision is replaced by internally learned, structure-aware reliability modeling.

Extensive experiments on both binary and multi-class datasets demonstrate that OFD-Net consistently improves segmentation performance under limited labeled supervision. The ablation studies further verify the contributions of OFDM, the orthogonal constraint, the reliability mechanism, dual-head fusion, and dual perturbation. In particular, the quartile analysis and qualitative results show that the proposed reliability mechanism not only improves the overall performance, but also functions in a way consistent with its design goal, namely, suppressing severe pseudo-label errors in structurally difficult cases.

Overall, the results suggest that explicit foreground-background structure can serve not only as a representation-learning prior, but also as a practical basis for reliable unlabeled supervision. We hope that this perspective can encourage further exploration of teacher-free and structure-aware semi-supervised segmentation methods in medical image analysis.

\section*{Declarations}
\label{sec:declarations}

\subsection*{Availability of data and materials}
\label{sec:data}
The ISIC-2016, Kvasir-SEG, Synapse, and ACDC datasets used in this study are all publicly available. All data generated and analyzed during this study are included in this published article.

\subsection*{Author Contribution}
\label{sec:Author Contribution}
All authors contributed to the conception and design of the study. Shao-feng Jiang led the methodology, drafted the initial manuscript, and secured funding. Zhe-yang Jing and Qin Lu implemented the software and conducted model training and validation. Huan-huan Shi, Zhen Chen, and Cong-xuan Zhang provided academic supervision and contributed to the study design and manuscript revision. Yi Chen supervised project progress and led the critical revision of the manuscript. All authors read and approved the final manuscript.

\subsection*{Ethics declarations}
\label{sec:ethics-declarations}
\noindent\textbf{Ethics approval and consent to participate}:
This study was approved by the Ethics Committee of Nanchang Hangkong University in China in accordance with the Declaration of Helsinki. This study is based on publicly available datasets and does not involve the direct participation of human or animal subjects. All data processing procedures comply with appropriate ethical standards and privacy protection.

\subsection*{Consent for publication}
\label{sec:Consent for publication}
Consent for publication is not applicable in this study, because there is not any individual person’s data.

\subsection*{Competing interests}
\label{sec:Competing interests}
There are no conflicts of interest to declare.

\subsection*{Acknowledgement}
\label{sec:Acknowledgement}
Not applicable.

\bibliographystyle{IEEEtran}
\bibliography{reference}

\vfill

\end{document}